\documentclass[letterpaper, 10 pt, conference]{ieeeconf}

\IEEEoverridecommandlockouts                              

\usepackage{myStyle}
\usepackage{siunitx}

\begin{document}

\title{\LARGE \bf
Interaction-Aware Sampling-Based MPC\\
with Learned Local Goal Predictions
}
\author{Walter Jansma, Elia Trevisan, Álvaro Serra-Gómez and Javier Alonso-Mora 
\thanks{This research is supported by the project ``Sustainable Transportation and Logistics over Water: Electrification, Automation and Optimization (TRiLOGy)'' of the Netherlands Organization for Scientific Research (NWO), domain Science (ENW), and the Amsterdam Institute for Advanced Metropolitan Solutions (AMS) in the Netherlands.}
\thanks{The authors are with the Cognitive Robotics Department,
       TU Delft, 
        {\tt\small walter.jansma@gmail.com}, {\tt\small \{e.trevisan, a.serragomez, j.alonsomora\}@tudelft.nl}}}
\maketitle
\thispagestyle{empty}
\pagestyle{empty}

\begin{abstract}
Motion planning for autonomous robots in tight, interaction-rich, and mixed human-robot environments is challenging. State-of-the-art methods typically separate prediction and planning, predicting other agents' trajectories first and then planning the ego agent's motion in the remaining free space. However, agents' lack of awareness of their influence on others can lead to the freezing robot problem.
We build upon Interaction-Aware Model Predictive Path Integral (IA-MPPI) control and combine it with learning-based trajectory predictions, thereby relaxing its reliance on communicated short-term goals for other agents.
We apply this framework to Autonomous Surface Vessels (ASVs) navigating urban canals. By generating an artificial dataset in real sections of Amsterdam's canals, adapting and training a prediction model for our domain, and proposing heuristics to extract local goals, we enable effective cooperation in planning. Our approach improves autonomous robot navigation in complex, crowded environments, with potential implications for multi-agent systems and human-robot interaction.\\
\newline
{Dataset, Prediction Model, Video and Code available at: \textcolor{cyan}{\rurl{autonomousrobots.nl/pubpage/IA_MPPI_LBM.html}}}

\end{abstract}


\section{Introduction}
Cities characterized by dense networks of urban canals, such as Amsterdam, could greatly benefit from deploying Autonomous Surface Vessels (ASVs) for various tasks including deliveries, transportation of people, and garbage collection \cite{wang_roboat_2020}.
However, navigating autonomously in urban canals amidst mixed human-robot crowds presents a significant challenge. Urban canals are typically narrow, frequently congested, and lack the structured nature of roads. While not as strictly enforced as on roads, navigation principles like right-of-way and right-hand conventions should still be considered.
Thus, akin to autonomous ground robots among pedestrian crowds, successful navigation in urban canals relies on cooperation and awareness of interactions \cite{trautman_unfreezing_2010}.

Recently, a sampling-based Model Predictive Control (MPC) called Interaction-Aware Model Predictive Path Integral (IA-MPPI) control has been developed for generating cooperative motion plans in urban canals among multiple non-communicating vessels while maintaining awareness of navigation rules \cite{streichenberg_multi-agent_2023}. This algorithm assumes rational and homogeneous agents, exact sensing of states, and knowledge of local goals. In real-time, the algorithm samples thousands of input sequences to approximate the optimal input sequence that enables all agents to progress toward their goals cooperatively.
In scenarios where the local goals of other vessels are unavailable, such as in mixed human-robot environments or due to lack of communication, this previous approach has approximated these goals using a constant velocity model over a given horizon. However, in narrow and crowded environments, vessels often need to execute complex maneuvers to navigate tight intersections and avoid collisions while adhering to navigation rules. In such situations, relying solely on a constant velocity approximation can lead to inaccurate predictions, which can adversely affect the performance of the motion planner in terms of deadlocks, collisions, navigation rule violations, traveled distance, and travel time.

\begin{figure}[t]
  \centering
  \includegraphics[width=0.95\linewidth]{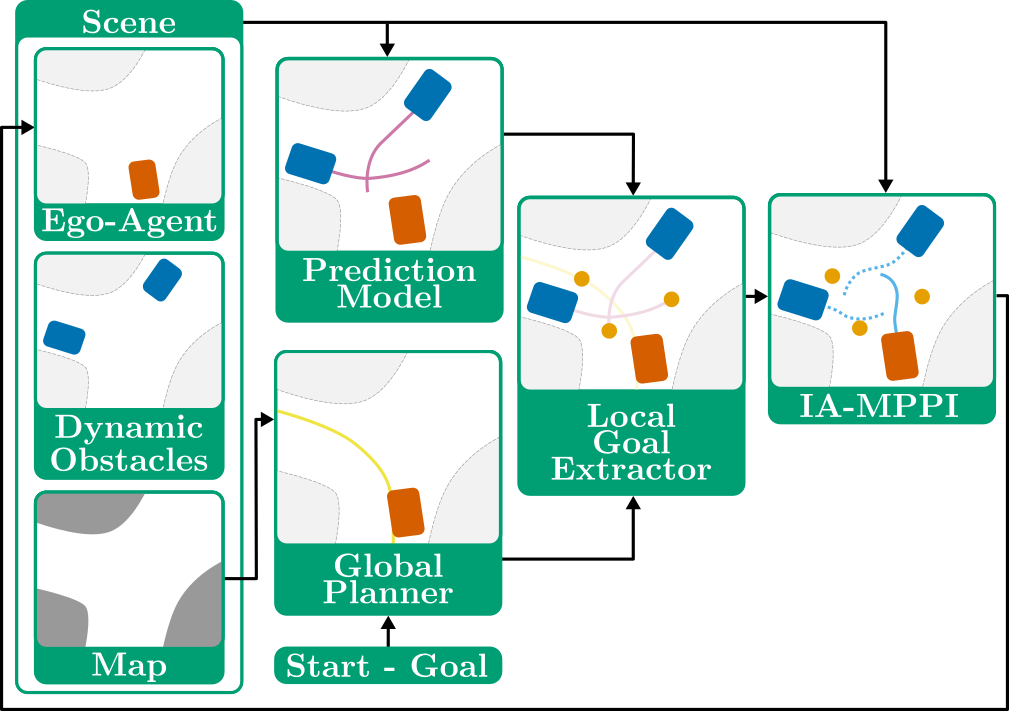}
  \caption{Overview of the proposed framework. Firstly, the prediction model utilizes information from all elements in the scene to forecast trajectories for obstacle agents. Meanwhile, the global planner, equipped with the map, start, and goal positions, generates a path for the ego agent. Subsequently, the local goal extractor leverages this information to determine appropriate local goals for the motion planner. With inputs derived from the scene and the local goals, the Interaction-Aware Model Predictive Path Integral (IA-MPPI) algorithm simultaneously plans and predicts input sequences for all agents in the scene. The first input of the sequence is then assigned to the ego agent, and the algorithm iterates.}
  \label{img:framework}
\end{figure}

In this paper, we present a framework (see Fig.~\ref{img:framework}) that utilizes a learning-based trajectory prediction method to improve the estimation of agents' intended destinations. We introduce heuristics to extract local goals from the predicted trajectories and provide the motion planner with the flexibility to influence the behavior of other agents while expecting cooperation in collision avoidance.

\subsection{Related Work}
Robot motion planning in dynamic environments is a challenging problem for which a series of classical and heuristic-based approaches have been developed~\cite{mohanan_survey_2018}, such as the Dynamic Window Approach ~\cite{fox_dynamic_1997} or Reciprocal Velocity Obstacles~\cite{van_den_berg_reciprocal_2008, van_den_berg_reciprocal_2011}. Despite their successful applications, e.g. to non-holonomic robots~\cite{alonso-mora_optimal_2013} or vessels in open waters~\cite{kuwata_safe_2014}, the motions planned by this class of methods are often reactive. This, especially in crowded environments, can lead to unsafe and unpredictable behaviors.

Model Predictive Control (MPC) has become a popular approach to trajectory planning for autonomous vehicles~\cite{zanon_model_2014} because of its ability to optimize accounting for the system's dynamics and constraints. Moreover, by planning over a sufficiently large horizon, MPC can anticipate dynamic obstacles resulting in trajectories that are less reactive.
To anticipate other agents, however, the free space over the entire planning horizon needs to be computed~\cite{Paden2016}, which requires knowledge about other agents' positions in the future.
If all the agents in the environment are autonomous, communication and distributed optimization can be used to plan trajectories in multi-agent environments~\cite{luis_online_2020}.
In mixed human-robot environments, however, such communication is not possible and predictions of the future motion of the other agents have to be employed. 
For instance, recent work on MPC for rule-aware navigation in urban canals uses constant velocity to model the future behavior of other vessels~\cite{de_vries_regulations_2022}.

In interaction-rich scenarios, however, constant velocity can be an inaccurate approximation which may lead to unsafe motion plans~\cite{Zhu}.
Therefore, several works rely on learning-based models to predict the future motion of other agents~\cite{zhu_learning_2021} and can include prediction confidence~\cite{fridovich-keil_confidence-aware_2020} and multimodality~\cite{schmerling_multimodal_2018}.
These methods, however, decouple prediction and planning which, in high-interaction environments, may lead the ego agent to wrongly assume that no collision-free path exists~\cite{mavrogiannis_core_2023}.
To avoid the so-called freezing robot problem the robot has to expect cooperation in collision avoidance from the other agents~\cite{trautman_robot_2015}.
Coupled prediction and planning can be done with MPC by modeling the interacting agents as a system, but it quickly becomes expensive to solve via constrained optimization leading to long computation times and short planning horizons ~\cite{liu_interaction-aware_2023, sadigh_planning_2016}.

Building upon a novel sampling-based Model Predictive Control (MPC) framework\cite{williams_information-theoretic_2018}, Interaction-Aware Model Predictive Path Integral (IA-MPPI) control \cite{streichenberg_multi-agent_2023} has successfully demonstrated decentralized coupled predictions and planning in real-time, accommodating long prediction horizons, nonlinear dynamics, and discontinuous cost functions in multi-agent environments. While IA-MPPI has exhibited superior performance compared to optimization-based MPC approaches that rely on fixed predictions of other agents' motion, it necessitates knowledge of their near-term local goals, which can either be communicated or estimated.

\subsection{Contribution}
This paper presents a novel framework for interaction-aware decentralized motion planning in urban canals without relying on communication. Our framework encompasses the following contributions:
\begin{itemize}
\item Realistic Dataset: We generate and publish a realistic dataset of simulated rule-abiding vessel trajectories in real sections of Amsterdam's urban canals.

\item Learning-Based Trajectory Prediction: We adapt a pedestrian prediction model \cite{brito_social-vrnn_2021} to vessels and train it specifically for urban canals. This approach enables us to generate trajectory predictions for other agents.

\item Local Goal Extraction: We propose heuristics to extract local goals from the predicted trajectories, thereby providing the motion planner with information about where agents intend to go.

\item Communication-Free Coupled Prediction and Planning: By combining the local goal extraction with the IA-MPPI control \cite{streichenberg_multi-agent_2023}, we achieve coupled prediction and planning without the need for communication. This approach ensures that the ego agent can influence the behavior of other agents while anticipating cooperation in collision avoidance.
    
\end{itemize}

We validate our planning framework through extensive simulated experiments, comparing it against baseline approaches and providing insights into the benefits of coupled prediction and planning over decoupled methods.
The framework can be adapted to other robot types beyond vessels.



\section{Interaction-Aware MPPI} \label{subsec:mppi}
In this section, we introduce the main ideas of IA-MPPI~\cite{streichenberg_multi-agent_2023}, upon which our proposed framework is built. For details on the method, models used and cost function please refer to the original paper. For insights on the underlying sampling-based MPC, one can refer to the work on Information-Theoretic MPC~\cite{williams_information-theoretic_2018}.
In short, IA-MPPI assumes that all the agents are homogenous and rational, i.e. have the same model and cost function. Under this assumption, we can create a large multi-agent system and plan input sequences resulting in cooperative trajectories for the ego agent as well as all the obstacle agents. This being a decentralized planning framework, we then apply the first input of the sequence to our ego agent, observe the environment and plan again.
In more detail, IA-MPPI models the ego-agent $i$ as a discrete-time dynamical system,
\begin{equation}
    \mathbf{q}_{i,t+1} = \mathcal{F}(\mathbf{q}_{i,t}, \mathbf{u}_{i,t})
\end{equation}
where $\mathbf{q}_{i,t}$ and $\mathbf{u}_{i,t}$ are, respectively, the state and the input of the ego-agent at timestep $t$. The state $\mathbf{q}_{i,t} = [\mathbf{p}_{i,t}, \mathbf{v}_{i,t}]$ contains the position and velocity of the agent.
IA-MPPI assumes that all agents in the environment are homogenous. The state and the input of the multi-agent system consisting of the ego-agent and the obstacle agents can therefore be stacked, resulting in,
\begin{equation}
\begin{aligned}
    \mathbf{q}&=\begin{bmatrix}\mathbf{q}_i^\top& \mathbf{q}^{\top}_j\end{bmatrix}^\top,\\
    \mathbf{u}&=\begin{bmatrix}\mathbf{u}_i^\top& \mathbf{u}^{\top}_j\end{bmatrix}^\top,
\end{aligned}
\quad\forall j\in\mathcal{M}\setminus i,
\end{equation}
where $\left(.\right)_j$ is a variable that the ego-agent $i$ estimates of agent $j$ and $\mathcal{M}=\{0,1,...,m\}$ is the set of all agents in the scene.
By also stacking the state transition functions $\mathcal{F}$ over all agents, we obtain a model for the multi-agent system $\mathbf{q}_{t+1}=\mathcal{G}(\mathbf{q}_t,\mathbf{u}_t)$.
Given a planning horizon $T$ and a prior input sequence $\mathbf{U} = [\mathbf{u}_0, \mathbf{u}_1,\dots, \mathbf{u}_{T-1}]$, IA-MPPI samples $K$ input sequences for the entire multi-agent system,
\begin{equation}
    \tilde{\mathbf{U}}_k = [\tilde{\mathbf{u}}_{0,k}, \tilde{\mathbf{u}}_{1,k}, \dots, \tilde{\mathbf{u}}_{T-1,k}], \quad \tilde{\mathbf{u}}_{t,k} = \mathcal{N}(\mathbf{u}_t, \nu \mathbf{\Sigma})
\end{equation}
with $k=1,\dots,K$, variance $\Sigma$ and scaling parameter $\nu$.
At the first iteration, the prior input sequence $\mathbf{U}$ is initialized at zero. By the end of this section, it will become clear how this prior input sequence is updated in subsequent iterations.
Having a model for the multi-agent system, we can forward simulate the $K$ input sequences into $K$ state trajectories $\mathbf{Q}_k$ for the multi-agent system,
\begin{equation}
\mathbf{Q}_k = \big[\mathbf{q}_0, \, \mathcal{G}(\mathbf{q}_0, \tilde{\mathbf{u}}_{k,0}), \, \dots, \, \mathcal{G}(\mathbf{q}_{k, T-1}, \,\tilde{\mathbf{u}}_{k,T-1})\big].
\end{equation}
Each of the resulting state trajectories is evaluated with respect to both an agent-centric cost as well as a system-wide cost, resulting in a total sample cost $S_k$.
The reader can refer to the original publication for details on the cost function~\cite{streichenberg_multi-agent_2023}. For the scope of our paper, it is important to know that the agent-centric cost includes a tracking cost to encourage progress towards a local goal $p_g$ computed as,
\begin{equation}\label{eq::ctracking}
    C_{\text{tracking}} = k_{\text{tracking}} \frac{\vert\vert\mathbf{p}_{g} - \mathbf{p}_{t}\vert \vert_2}
    {\vert\vert\mathbf{p}_{g} - \mathbf{p}_{t_0}\vert \vert_2},
\end{equation}
where $p_t$ is the position of the agent at timestep $t$, $p_{t_0}$ is the position of the agent at the beginning of the planning horizon and $k_{tracking}$ is a tuning parameter.
Notice that we need to know the position of the local goal of each agent. For the ego agent, the local goal is extracted from a global plan. For all the other agents, the local goal has to be either communicated or estimated.
We propose in the following section how this goal can be estimated.
Once $S_k$, $\forall k \in [1,\dots, K]$ has been computed, importance sampling weights $w_k$ can be calculated as,
\begin{equation}
    w_k = \frac{1}{\eta} \exp\biggl( \frac{-1}{\lambda}(S_k - S_{min})\biggr), \quad \sum_{k = 0}^{K-1} w_k  = 1, 
\end{equation}
where $S_{min}$ is the minimum sampled cost, $\eta$ a normalization factor and $\lambda$ a tuning parameter.
We then compute an approximation of the optimal control sequence through a weighted average of the sampled control sequences,
\begin{equation}
    \mathbf{U}^* =  \sum_{k = 0}^{K-1} w_k \tilde{\mathbf{U}}_k
\end{equation}
and apply the first input $\mathbf{u}^*_{i,0}$ to the ego-agent. We can now use a time-shifted version of $\mathbf{U}^*$ as the prior input sequence $\mathbf{U}$ to warm-start the sampling strategy at the next iteration.

\section{Predicting goal positions} 
In Fig.~\ref{img:framework} we provide an overview of the proposed framework. 
In Section~\ref{subsec:socialvrnn}, we outline the prediction model.
In Section~\ref{subsec:data}, we describe the dataset we have collected to train a prediction model that is interaction and rule-aware. In Section~\ref{subsec:training}, we present the steps taken to port the prediction model to urban vessel environments. In Section~\ref{subsec:heuristics}, we propose a heuristic to extract a local goal suitable for IA-MPPI using the predicted trajectories.

\subsection{Interaction-aware trajectory prediction method}
\label{subsec:socialvrnn}
Our approach leverages interaction-aware trajectory prediction for goal estimation. We employ an adapted version of \textit{Social-VRNN}~\cite{brito_social-vrnn_2021}, which was originally designed for pedestrians, to obtain trajectory predictions. However, we remark that our framework is agnostic to the choice of trajectory predictor as long as it accounts for obstacles and interactions between agents in the environment.

Social-VRNN \cite{brito_social-vrnn_2021} is an interaction-aware trajectory prediction method that leverages a generative model based on Variational Recurrent Neural Networks (VRNNs) \cite{chung_recurrent_2015}. 
The model combines three types of contextual cues to define a joint representation of an agent's current state: information on the past trajectory of the agent of interest, environment context, and agent-agent interactions. The input to predict the trajectory of agent $i$ is denoted as:
\begin{equation}
    \textbf{x} = \{\textbf{v}^{i}_{-T_{0}:0}, \textbf{O}^{i}_{env},\textbf{O}^{-i}_{int}\},
\end{equation}
where $\textbf{v}^{i}_{-T_{0}:0}$ corresponds to the sequence of velocity states over the previous observed horizon $T_{O}$ of the agent of interest $i$.
The environment information $\textbf{O}^{i}_{env}$ is represented in the form of a grid map extracted around the agent of interest. Then, $\textbf{O}^{-i}_{int}$ represents the information on agent-agent interactions. It is a vector with the relative positions and velocities of all other agents from agent $i$'s perspective, listed in ascending order based on the absolute distance to it. The output of the model is a sequence of velocity probability distributions represented by $T_H$ diagonal gaussian distributions $\mathcal{N}(\mu_{\textbf{v}, k}, \text{diag}(\sigma^{2}_{\textbf{v},k}))$. For details on the method and its architecture, please refer to the original paper \cite{brito_social-vrnn_2021}.

\subsection{Artificial Dataset} \label{subsec:data}
\begin{figure}[ht]
  \centering
  \includegraphics[width=1.05\linewidth]{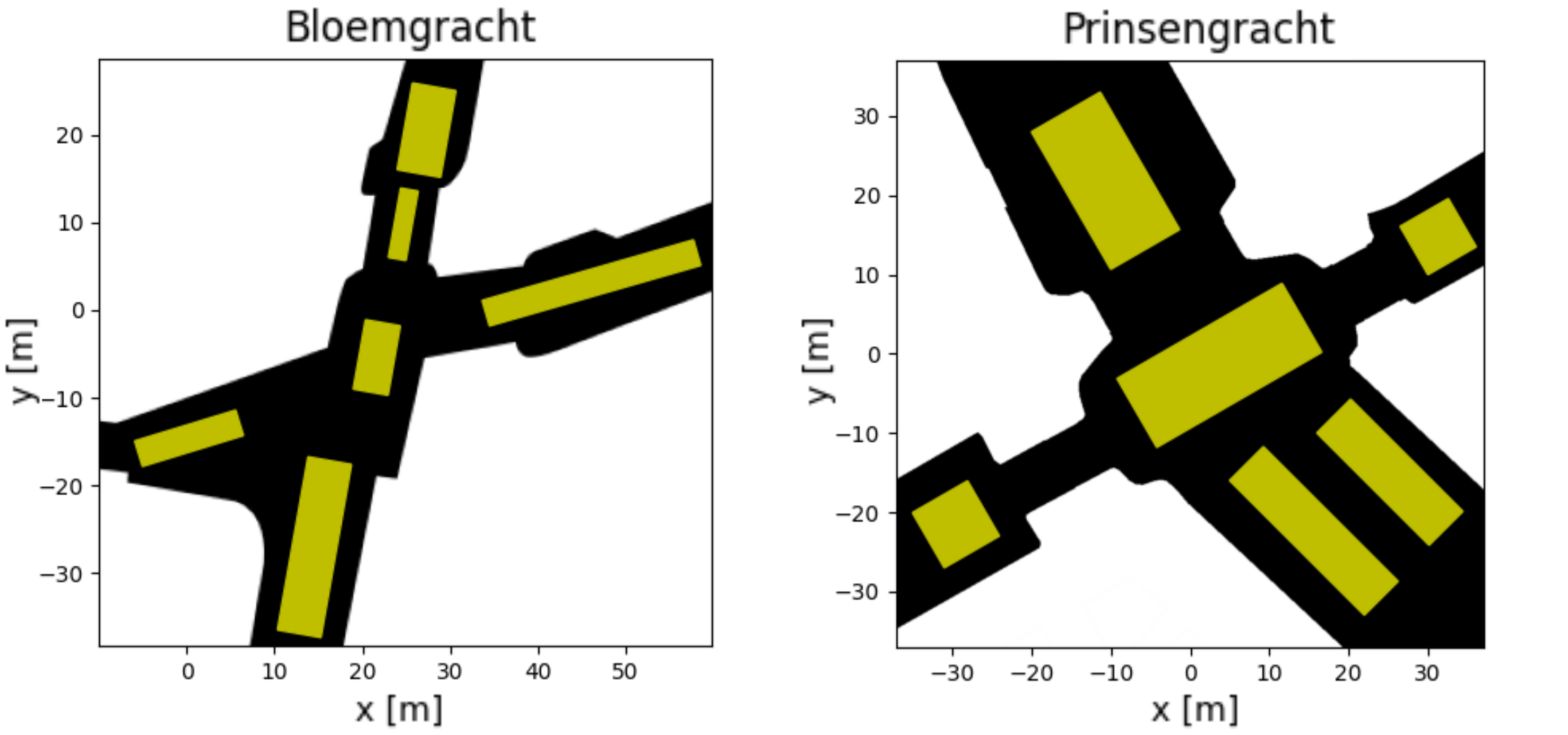}
  \caption{Canal sections of the Bloemgracht and Prinsengracht. The black areas are the canals. The yellow rectangles correspond to the initialization areas in which goals and starting locations were randomly initialized for each agent during the simulations.}
  \label{fig:data}
\end{figure}

In the absence of a publicly available dataset for short-term vessel trajectory prediction, an artificial dataset of vessel interactions is collected in a simulation environment. In order to obtain trajectories that resemble those of real vessels in urban canals, four real canal section maps in Amsterdam: the Herengracht (HG), the Prinsengracht (PG) and the Bloemgracht (BG) are used to collect data. The Open Crossing (OC) environment is created to collect vessel interactions in open water. Data on an additional environment, the Amstel (AM), is included only for testing our framework's generalization to environments not seen during training.  Figure \ref{fig:data} depicts two of these canal sections. The yellow rectangles correspond to the areas in which start and goal locations are randomly initialized. These areas are placed around the entire map and in each canal section to improve the diversity of the trajectories and interactions.

To collect the data, more than four thousand experiments are conducted by initializing up to four vessels simultaneously in the mentioned environments. Each vessel is assigned a randomized start and goal location in one of the predefined areas. All sampled locations are ensured to be collision-free. The vessels run a centralized IA-MPPI to sail toward their respective goals while accounting for navigation rules. This ensures that the recorded trajectories are safe, interaction-aware, and mostly rule-abiding.

For each experiment and vessel in the environment we record the current timestamp, the vessel ID, its position and velocity in the global frame. Each timestamp is unique across timesteps and experiments, which enables to identify vessels belonging to the same scene.
The specifications of the artificial vessel dataset can be found in Table \ref{tab:data}. In order to evaluate the prediction model, 10\% of the dataset is used as the test set. The remaining data is used for training and is split into a training set (72\%) and a validation set (18\%). The distribution of data from each scenario is ensured to be equal in all splits. 

\begin{table}
    \medskip
\centering
  \caption{Specifications of the artificial dataset. \textit{Exp.} refers to the number of experiments done in each scenario. \textit{Frames} and \textit{Vessels} refer to the total number of frames and vessels present in the data set, respectively. All data is recorded at a rate of \qty{10}{\hertz}.}
  \label{tab:data}
  \begin{tabular}{rlll}\toprule
    \textbf{Scenario} & \textbf{Exp.} & \textbf{Frames} & \textbf{Vessels} \\ \midrule
    Herengracht & 1000 & 406229 & 2499 \\
    Prinsengracht & 1247 & 420285 & 4122 \\
    Bloemgracht & 1188 & 372173 & 3564\\
    Open Crossing & 1182 & 417515 & 3544  \\ 
    Amstel & 79 & 23468 & 316 \\
    \textbf{Total} & 4696 &\textbf{1639670} & \textbf{14045}\\
    \bottomrule
  \end{tabular}
\end{table}

\subsection{Model Training and Adaptation} \label{subsec:training}
We adapt the variational inference architecture presented in \cite{brito_social-vrnn_2021} to generate unimodal trajectory probability predictions of vessels. In contrast to humans, vessels are slower and have lower-order dynamics, which results in less reactive behaviors and smoother trajectories. To take this into account and avoid overfitting to the dataset, we reduce the dimensionality of the method's latent space. We also add an L2-regularization term to the loss function and weight it with a hyperparameter we define as $\gamma$.

\subsubsection{Hyperparameters}
The model is trained using backpropagation through time and the RMSProp \cite{tieleman_lecture_2012} optimizer. With a time step of $\Delta T = 0.4$ seconds, the prediction horizon is set to $T_H = 24$ steps (9.6 seconds) and the previous horizon to $T_O = 14$ steps (5.6 seconds). Furthermore, we employ learning rate starting at $\alpha$ = $1\mathrm{e}{-4}$ that decays by a factor of 0.9 after every gradient step. The regularization weight is kept at $\gamma = 0.0001$. Finally, the model is trained for $4\mathrm{e}{4}$ training steps, using early stopping.

\subsection{Local Goal Extraction} \label{subsec:heuristics}
\begin{figure}[h]
  \centering
  \medskip
  \includegraphics[width=1\linewidth]{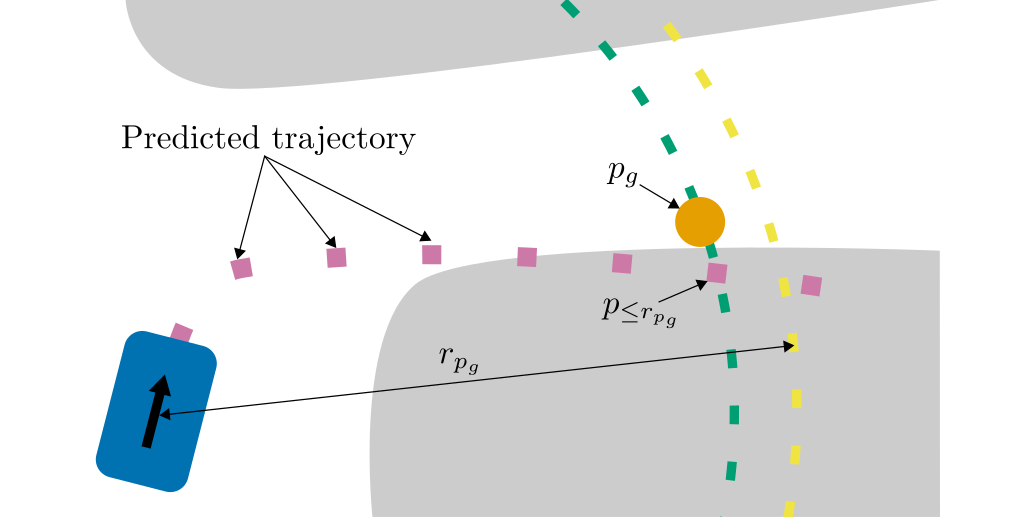}
  \caption{A visual illustration of how the local goal is extracted from a colliding trajectory prediction.}
  \label{fig:localgoal}
\end{figure}

In eq.~\eqref{eq::ctracking} we show that the IA-MPPI needs to know the local goal $p_g$ of each agent.
There are two requirements for a goal to be suitable: it has to lie within a radius $r_{p_g}$ from the agent it corresponds to and cannot be in space occupied by static obstacles.
Therefore, we first search the predicted trajectory backward until we obtain a position $p_{\leq r_{p_g}}$ within the desired radius.
If $p_{\leq r_{p_g}}$ is in collision with a static obstacle, we construct a circle centered on the agent's position $p_a$ with radius $p_a-p_{\leq r_{p_g}}$ and find the point on the circle closest to $p_{\leq r_{p_g}}$ which is not in collision with static obstacles. This goal extraction method is illustrated in Fig.~\ref{fig:localgoal}.
Once the goals for all agents are predicted, IA-MPPI can plan interaction-aware trajectories in a decentralized fashion.

\begin{figure*}[ht!]

    \medskip
  \centering
  \includegraphics[width=\textwidth]{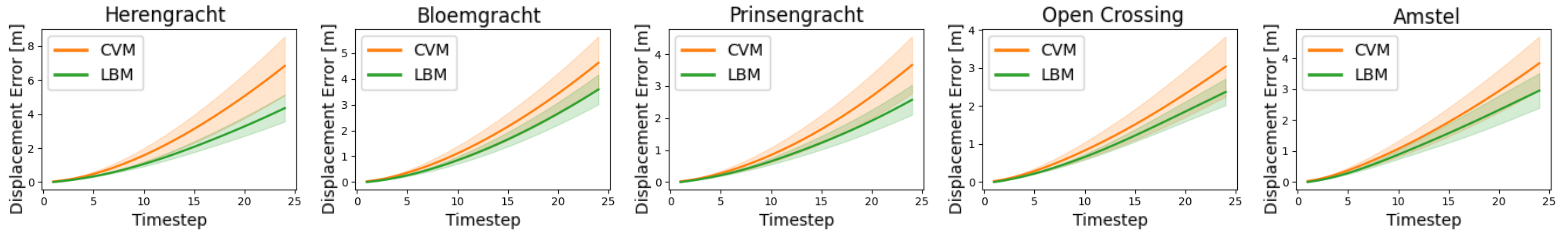}
  \caption{The displacement error of the predictions from CVM and the LBM over the prediction horizon for each canal section. The solid line represents the mean error and the shaded area represents 30\% of the standard deviation.}
  \label{fig:fde}
\end{figure*}

\section{Experiments}
The experiments are conducted in real maps of Amsterdam's canals, namely the Herengracht (HG), Bloemgracht (BG), Prinsengracht (PG), and the Amstel (AM). In addition, experiments are conducted in an Open Crossing (OC) map without static obstacles. In Section \ref{subsec:fde} we evaluate the prediction model, in Section \ref{subsec:decIAMPPI} we show the performances of the proposed framework for motion planning, and in Section \ref{subsec:noninteractive} we highlight the benefits of coupled prediction and planning with respect to a decoupled approach.

\subsection{Prediction Accuracy} \label{subsec:fde}
In Fig.~\ref{fig:fde} we compare the proposed Learning-Based Model (LBM) to a Constant Velocity Model (CVM) on test data. We evaluate the methods against the displacement error at each prediction step, which is defined as the Euclidean distance between a prediction and the ground truth.
In all maps the LBM outperforms the CVM, showing a lower average displacement error and a smaller standard deviation.
Note that the Amstel map was previously unseen during training, demonstrating generalization capabilities.

\begin{figure}[ht]
  \centering
  \includegraphics[width=0.45\textwidth]{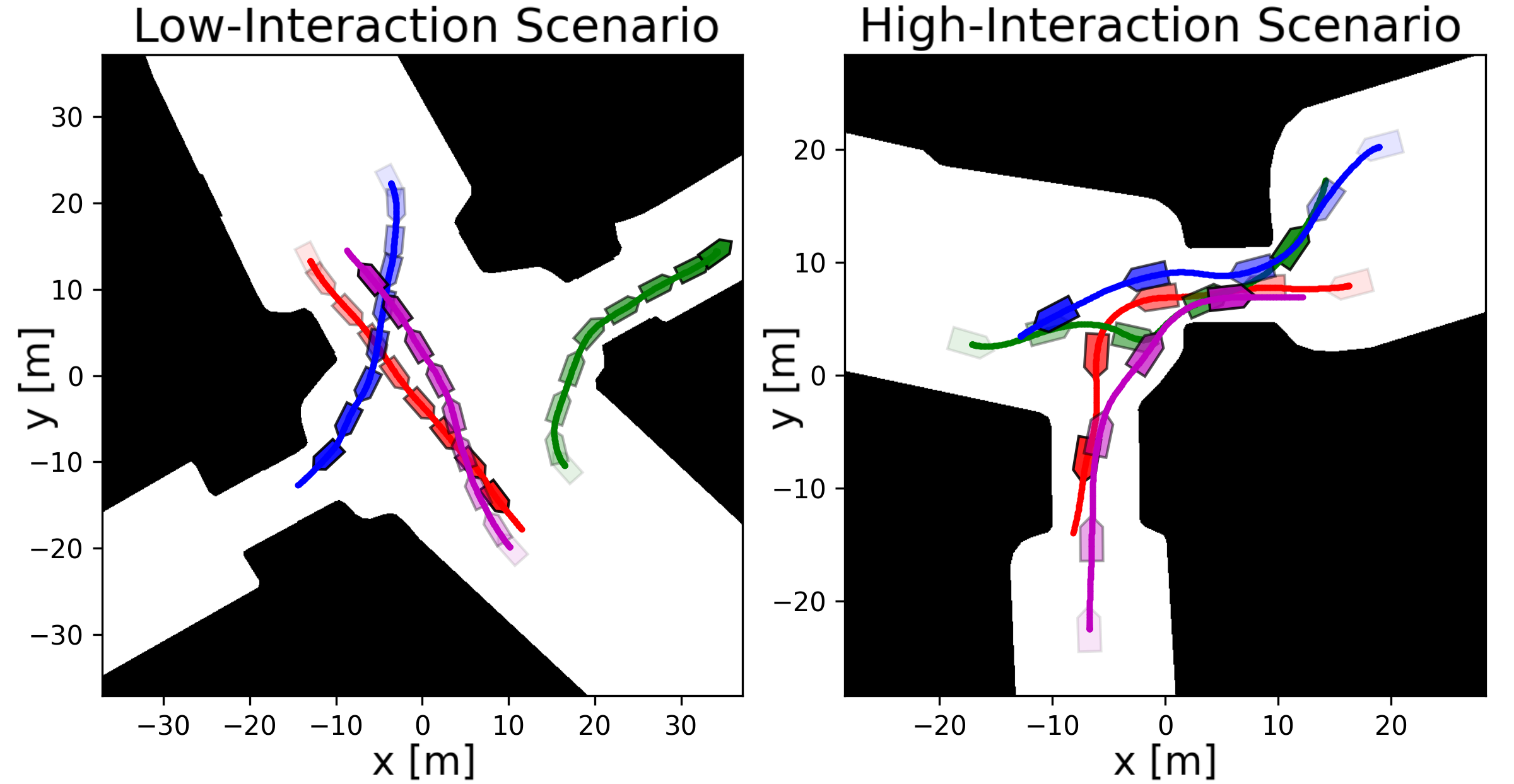}
  \caption{Examples of experiments in the low-interaction scenario (left) and high-interaction scenario (right).}
  \label{fig:scenarios}
\end{figure}

\subsection{Interaction-Aware Motion Planning with Predictions}
\label{subsec:decIAMPPI}
In this study, we evaluate the performance of the proposed decentralized framework that uses a Learning-Based prediction Model to extract local goals (IA-MPPI-LBM), by comparing it against a decentralized approach that extracts local goals from a Constant Velocity Model (IA-MPPI-CVM) and decentralized with communication (IA-MPPI-w/comm.), which assumes perfect knowledge of other agents' local goals.
It is important to stress that, in similar experiments, the IA-MPPI-CVM which serves as the communication-free baseline in our comparisons has already been demonstrated to outperform an optimization-based Model Predictive Control (MPC) approach that relies on fixed predictions~\cite{streichenberg_multi-agent_2023}.

In the simulated experiments taking place in real sections of the canals of Amsterdam, we randomize the initial positions and goals of four interacting agents, all running the same algorithm.
To challenge each method, we design regions within which each agent's start and goal position are randomly initialized in a way that forces all four agents to interact in a narrow section of the map. These \textit{high-interaction} scenarios are discussed in Section~\ref{highinter}.

For completeness, we also design experiments where agents' starting and goal positions are randomized across much larger spaces. In these experiments, however, vessels don't often interact and usually have larger free spaces to avoid each other. These \textit{low-interaction} scenarios are discussed in Section~\ref{lowinter}.

An example of experiments in low- and high-interaction scenarios is shown in Fig.~\ref{fig:scenarios}. 
The IA-MPPI plans with a time horizon $T$ of 100 time steps with step size $\delta T = 0.1s$ and $K=4500$ samples. Each method is evaluated on the same set of randomly initialized experiments. For fairness, metrics such as rule violations, goal displacement error, total traveled distance, and time are only displayed for experiments that ended successfully with all methods.

\begin{table}[ht]
  \caption{Successes (Succ.), Deadlocks (Deadl.), Collisions (Coll.), Rule Violations (Rule Viol.) and Goal Displacement Error (Goal DE) for all methods in high-interaction scenarios per canal sections.}
  \label{tab:complexinteractive}
  \resizebox{\columnwidth}{!}{
  \begin{tabular}{clcccc}\toprule
    & \multirow{2}{*}{\textbf{Method}} & \textbf{Succ. / Deadl.} & \multirow{2}{*}{\textbf{Rule Viol.}} & \multirow{2}{*}{\textbf{Goal DE}} \\ 
    & & \textbf{/ Coll.} \cr \midrule
    
    \multirow{4}{*}{\rotatebox{90}{HG}}   
    & IA-MPPI-CVM & 18 / 0 / 2 & 16 & 5.82 m\\
    & IA-MPPI-LBM (ours) & 19 / 1 / 0 & 16 & 5.26 m\\ 
    \cmidrule{2-5}
    & IA-MPPI-w/comm. & 20 / 0 / 0 & 16 \cr \midrule

    \multirow{4}{*}{\rotatebox{90}{PG}}
    & IA-MPPI-CVM & 19 / 0 / 1 & 11 & 7.30 m\\
    & IA-MPPI-LBM (ours) & 19 / 1 / 0 & 5 & 4.11 m\\
    \cmidrule{2-5}
    & IA-MPPI-w/comm. & 20 / 0 / 0 & 5 \cr \midrule

    \multirow{4}{*}{\rotatebox{90}{BG}}
    & IA-MPPI-CVM & 17 / 0 / 3 & 7 & 4.98 m \\
    & IA-MPPI-LBM (ours) & 20 / 0 / 0 & 4 & 4.13 m\\ 
    \cmidrule{2-5}
    & IA-MPPI-w/comm. & 20 / 0 / 0 & 3 \cr \midrule
  \end{tabular}
}
\end{table}

\subsubsection{High-Interaction Scenario} \label{highinter}
\begin{figure}[h]
  \centering
  \includegraphics[width=0.45\textwidth]{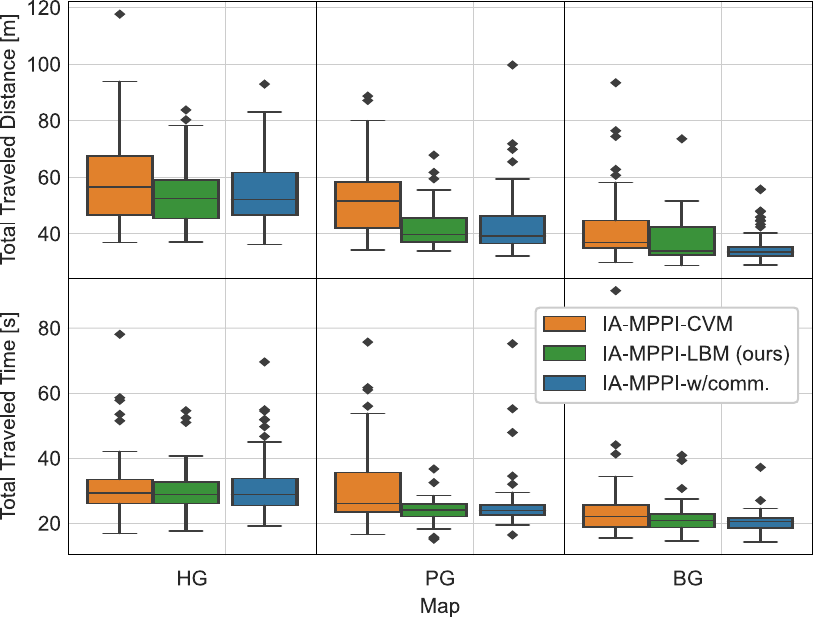}
  \caption{This figure displays the distribution of the total traveled distance and total traveled time of the vessels during the experiments in the high-interaction scenarios. The results are displayed per map and for each method.}
  \label{fig:complexdistance}
\end{figure}

The experiments in high-interaction scenarios are conducted in narrow intersections in the Bloemgracht, Herengracht, and Prinsengracht. Since the Amstel canal is very wide and the Open Crossing has no static map constraints, it is difficult to generate experiments with high-interactions, and thus these two maps are excluded from this experiment section. The results of the experiments are summarized in Table \ref{tab:complexinteractive} and Figure \ref{fig:complexdistance}.
It can be seen that in these high-interaction scenarios, the LBM consistently outperforms the CVM in terms of the goal displacement error (Goal DE).
As a consequence, the motion planning framework that estimates other agents' local goals using predictions from the LBM outperforms the framework that uses the CVM on all the metrics. 
Moreover, we demonstrate our framework with the LBM has negligible performance losses compared to the method with perfect communication.

\subsubsection{Low-Interaction Scenarios}\label{lowinter}
\begin{figure}[h]
  \centering
  \medskip
  \includegraphics[width=0.45\textwidth]{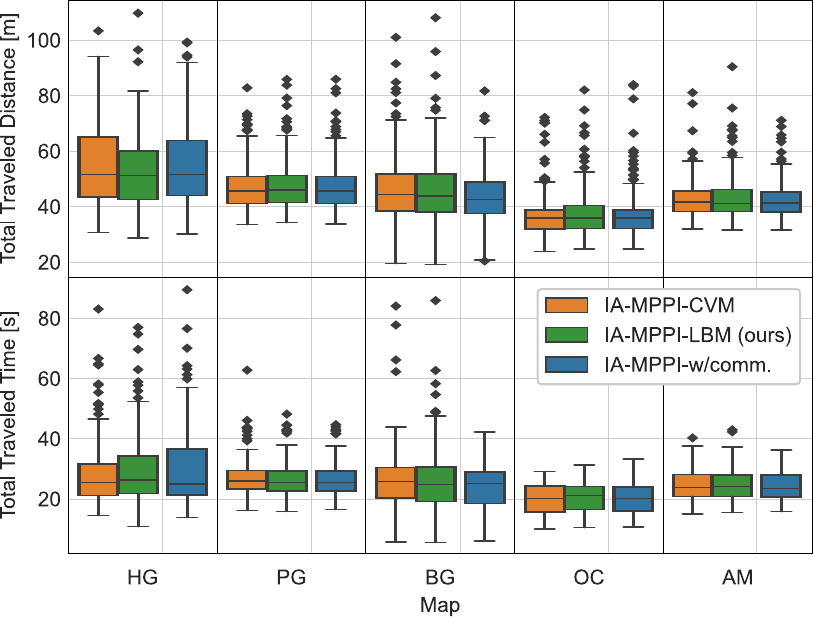}
  \caption{This figure displays the distribution of the total traveled distance and total traveled time by vessels during the experiments in the low-interaction scenarios. The results are displayed per map and for each method.}
  \label{fig:distance}
\end{figure}

Table~\ref{tab:interactive} and Fig.~\ref{fig:distance} summarize the results in low-interaction scenarios.
Note that we here also test on the Open Crossing maps and the Amstel, which the LBM has not previously seen in training.
The results show that also when the start and goal positions of all agents are randomly initialized over large areas, our proposed communication-free framework with the LBM performs just as well as the baseline with full communication, even in a map unseen in training.
However, perhaps unsurprisingly, the framework that approximates the local goals with a CVM can also achieve the same performance as the framework with full communication. Intuitively, in low-interaction scenarios where agents mostly navigate straight to their goal, CVM is a reasonably good approximator.


\subsection{Decoupled Prediction and Planning}
\label{subsec:noninteractive}
The framework we proposed utilizes a Learning-Based Model (LBM) to predict trajectories for obstacle agents and extract local goals while employing Interaction-Aware Model Predictive Path Integral (IA-MPPI) for coupled predictions and planning.
To assess the advantages of this framework, we compare it to a planner without interaction awareness (MPPI-LBM), which decouples prediction and planning. Like other state-of-the-art methods, MPPI-LBM treats the predicted future trajectories of obstacle agents as occupied space and plans the ego agent's motion without considering interaction awareness.
This approach reduces the system size and computational burden by minimizing the space to be sampled. However, apart from this difference, MPPI-LBM shares the same sampling strategy and cost function as the proposed IA-MPPI-LBM.
We conducted 100 low-interaction experiments across the Amstel, Bloemgracht, Herengracht, Open Crossing, and Prinsengracht, comparing different methods. Table \ref{tab:noninteractive} presents the outcomes, including total successes, deadlocks, collisions, and rule violations.
Again, the proposed IA-MPPI-LBM shows similar performances to the method with communication (IA-MPPI-w/comm).

However, MPPI-LBM exhibited a significantly lower success rate and a higher number of rule violations. The LBM, while trained to be somewhat rule- and interaction-aware in its predictions, occasionally struggles to capture complex reciprocal collision avoidance maneuvers when agents are in close proximity. This, combined with the motion planner's unawareness of the ego agent's influence on other agents' motion and their cooperation in collision avoidance, often led the MPPI-LBM to wrongly assume that no feasible solution existed. Consequently, this resulted in agents drifting into collisions due to their large inertia.

\begin{table}[t]

    \medskip
  \caption{Successes (Succ.), Deadlocks (Deadl.), Collisions (Coll.), Rule Violations (Rule Viol.), and Goal Displacement Error (Goal DE) for all methods in the various canal sections.}
  \label{tab:interactive}
  \resizebox{\columnwidth}{!}{
  \begin{tabular}{clcccc}\toprule
    & \multirow{2}{*}{\textbf{Method}} & \textbf{Succ. / Deadl.} & \multirow{2}{*}{\textbf{Rule Viol.}} & \multirow{2}{*}{\textbf{Goal DE}} \\ 
    & & \textbf{/ Coll.} \cr \midrule
    
    \multirow{4}{*}{\rotatebox{90}{HG}}
    & IA-MPPI-CVM & 38 / 0 / 2 & 22 & 5.31 m\\
    & IA-MPPI-LBM (ours) & 37 / 0 / 3 & 26 & 5.42 m\\
    \cmidrule{2-5}
    & IA-MPPI-w/comm. & 37 / 2 / 1 & 26 \cr \midrule

    \multirow{4}{*}{\rotatebox{90}{PG}}
    & IA-MPPI-CVM & 38 / 0 / 2 & 14 & 3.12 m \\
    & IA-MPPI-LBM (ours) & 38 / 0 / 2 & 13 & 2.94 m\\
    \cmidrule{2-5}
    & IA-MPPI-w/comm. & 38 / 0 / 2 & 14 \cr \midrule

    \multirow{4}{*}{\rotatebox{90}{BG}}
    & IA-MPPI-CVM & 38 / 0 / 2 & 17 & 5.36 m\\
    & IA-MPPI-LBM (ours) & 39 / 0 / 1 & 20 & 4.93 m\\
    \cmidrule{2-5}
    & IA-MPPI-w/comm. & 40 / 0 / 0 & 20\cr \midrule

    \multirow{4}{*}{\rotatebox{90}{OC}}
    & IA-MPPI-CVM & 40 / 0 / 0 & 27 & 3.98 m\\
    & IA-MPPI-LBM (ours) & 40 / 0 / 0 & 29 & 4.04 m\\
    \cmidrule{2-5}
    & IA-MPPI-w/comm. & 40 / 0 / 0 & 27 \cr \midrule

    \multirow{4}{*}{\rotatebox{90}{AM}}
    & IA-MPPI-CVM & 40 / 0 / 0 & 22 & 3.12 m\\
    & IA-MPPI-LBM (ours) & 39 / 1 / 0 & 18 & 3.49 m\\
    \cmidrule{2-5}
    & IA-MPPI-w/comm. & 40 / 0 / 0 & 21 \cr \midrule

  \end{tabular}
  }
\end{table}

\begin{table}[h]
  \caption{Successes (Succ.), Deadlocks (Deadl.), Collisions (Coll.), and Rule Violations (Rule Viol.) for the non-interactive MPPI and the IA-MPPI baseline in low-interaction scenarios.} 
  \label{tab:noninteractive}
  \resizebox{\columnwidth}{!}{
  \begin{tabular}{lcc}\toprule
    \textbf{Method} & \textbf{Succ. / Deadl. / Coll.} & \textbf{Rule Viol.} \\ \midrule
    MPPI-LBM. & 72 / 1 / 27 & 40 \\ 
    IA-MPPI-LBM (ours) & 97 / 1 / 2 & 30 \\
    IA-MPPI-w/comm. & 98 / 0 / 2 & 28 \\ \midrule
  \end{tabular}
  }
\end{table}

\section{Conclusions}

In this paper, we introduced a framework that combines a learning-based trajectory prediction model with Interaction Aware MPPI, enabling decentralized and communication-free coupled prediction and planning. Our experimental results demonstrated the superiority of our Learning-Based Model (LBM) over the Constant Velocity Model (CVM) in accurately predicting the trajectories of interacting vessels, even in unseen maps. Through simulated experiments in Amsterdam's canals, we showed that our motion planning framework achieved comparable performance to a method with ground truth knowledge of local goals, which was shown to outperform classical optimization-based MPC approaches with decoupled prediction and planning in previous work~\cite{streichenberg_multi-agent_2023}. Additionally, we highlighted the limitations of the CVM in tight environments with multiple interacting agents. Finally, by comparing our approach with a non-interactive planner, we emphasized the advantages of coupled planning and predictions.


\bibliographystyle{IEEEtran}
\bibliography{IEEEabrv,mybib}



\end{document}